# FERI: A Multitask-based Fairness Achieving Algorithm with Applications to Fair Organ Transplantation

Can Li, MS[1], Dejian Lai, PhD[1], Xiaoqian Jiang, PhD[2], Kai Zhang, PhD[2, *]
[1]Department of Biostatistics and Data Science, School of Public Health, The University of Texas Health Science Center at Houston, Houston, TX, USA
[2]Department of Health Data Science and Artificial Intelligence, McWilliams School of Biomedical Informatics, The University of Texas Health Science Center at Houston, Houston, TX, USA
*Corresponding Author: Kai.Zhang.1@uth.tmc.edu

**Abstract**

*Liver transplantation often faces fairness challenges across subgroups defined by sensitive attributes such as age group, gender, and race/ethnicity. Machine learning models for outcome prediction can introduce additional biases. Therefore, we introduce **F**airness through the **E**quitable **R**ate of **I**mprovement in Multitask Learning (FERI) algorithm for fair predictions of graft failure risk in liver transplant patients. FERI constrains subgroup loss by balancing learning rates and preventing subgroup dominance in the training process. Our results show that FERI maintained high predictive accuracy with AUROC and AUPRC comparable to baseline models. More importantly, FERI demonstrated an ability to improve fairness without sacrificing accuracy. Specifically, for the gender, FERI reduced the demographic parity disparity by 71.74%, and for the age group, it decreased the equalized odds disparity by 40.46%. Therefore, the FERI algorithm advanced fairness-aware predictive modeling in healthcare and provides an invaluable tool for equitable healthcare systems.*

## 1. Introduction

Deep neural networks have revolutionized several areas within the medical field, including medical image processing[1], patient risk prediction[2], and the interpretation of clinical notes[3]. These advancements have enhanced personalized medicine and the development of effective treatments for various diseases[4]. The efficacy of deep learning models can be partly attributed to data-driven representation learning, which enables automatic extraction of useful features directly from data[5]. However, real-world data often encodes historical prejudices and demographic inequalities[6]. Therefore, reliance on potentially biased data risks amplification of unfairness in trained models. The issue of fairness in healthcare, in particular, has been a long-standing topic of discussion in various literature[7,8]. The essence of fairness in healthcare requires addressing health disparities and reducing inequalities in both access to care and health resource allocation[9]. It requires implementing policy interventions targeted at underrepresented populations to bridge gaps in care. In recent years, attention has come to the fairness issues in healthcare prediction models[10]. Prediction models often emphasize accuracy enhancement without explicitly addressing issues of unfairness, potentially leading to inequitable outcomes.

Unfairness can originate from biases introduced in different stages of the machine learning pipeline, from data collection to model training. The source of bias often determines the approach used to achieve fairness. Based on where bias is addressed in the deep learning pipeline, fairness algorithms can be classified as pre-processing, in-processing, or post-processing methods[11]. Pre-processing algorithms focus on addressing biases before modeling, i.e., by processing the input data to mitigate biases that can potentially cause unfairness in the prediction outcomes[12]. Post-processing methods adjust model outputs after training to align predictions with fairness criteria. Examples include calibrated equalized odds for recalibrating predictions[13] and a fairness-aware ensemble framework to shift decision boundaries[14].

The goal of in-processing methods is to achieve model fairness during model training, and the critical approach is to seek strategies to balance model accuracy and model fairness[15,16]. It can usually be done in various ways, for example, conventional methods using constraint optimization to add the fairness metric as a regularization term in the optimization target function[6]. The fairness can be modeled by the model performance disparity among different subgroups. Therefore, it can be easily expressed as the loss difference among subgroups[17,18]. Several research studies have focused on directly modeling the fairness metrics as regularization terms[19]; however, the main difficulty is that most fairness metrics are "rate-based" expressions, which jeopardize the differentiability of the loss function[20]. Typical methods to alleviate this is to use differentiable functions as surrogate functions[21]. Furthermore, alternative methodologies have been proposed. Agarwal et al. used game theory to convert constrained optimization into a gameplay problem with two players competing two-players[22]; one optimizes the utility, and the other optimizes fairness. Kumar et al. treated the threshold as a learnable and differentiable parameter, making the rate-based fairness metrics a continuous variable[23]. In recent years, other novel in-processing methods have been proposed beyond constrained optimization to achieve model fairness, such as adversarial learning[24] and implicit dynamic reweighting[25].

This paper uses a multitask learning framework to explore in-processing techniques for achieving group fairness in machine learning models. Achieving fairness in machine learning often requires careful calibration of model performance across different subgroups, especially those defined by sensitive attributes such as gender, age, race, and ethnicity. This goal of balanced performance aligns with the multitask learning framework, where a model can concurrently learn multiple tasks[26]. The rationale for this analogy is on the concerns of the training speed disparity problem essentially[27]. Techniques from multitask learning, including dynamic weighting scheme and specialized regularization, aim to equalize various learning speeds inherently[28]. Meanwhile, applying such methods to fairness could also help address the disparity of the training speed between subgroups.

In this work, we introduce a novel algorithm, *Fairness through Equitable Rate of Improvement* (FERI), inspired by the *Fast Adaptive Multitask Optimization* (FAMO) to equalize subgroup training speeds[29]. Our key insight in FERI is that the synchronized optimization framework of FAMO can be adapted to address unfairness in machine learning. We propose constraints on each subgroup loss to equalize the rate of decrease and smooth the learning trajectory, preventing any one subgroup from dominating the learning process. We build on multitask learning concepts to address fairness challenges in this novel way. Our technique promotes fairness by balancing subgroup optimization rather than directly regulating model parameters. Through experiments, we validate that this FAMO-inspired technique aligns with and even improves upon current methods by improving equity and mitigating bias.

Liver transplantation is a critical medical procedure that exhibits systemic inequalities related to gender, age group, and racial and ethnic backgrounds[30]. For instance, Black patients are less likely to receive a liver transplant compared to white patients, highlighting racial disparities in access[31]. Similarly, there has been an increasing disparity in transplant rates between females and males[32]. The MELD score uses creatinine levels as one of the standards for liver transplant allocation, which are generally lower in women, leading to longer wait times and decreased access to liver transplantation for women[33]. Age-related disparities are also evident, as pediatric candidates face challenges due to the scarcity of suitably sized organs. While previous papers concentrate on the multiple post-liver transplantation risk factors predictions[34,35] there is also work by Ding et al. that predicts graft failure in a distinct approach to address fairness challenges in liver transplantation[36]. This paper focuses on these fairness challenges in liver transplantation, aiming to develop a more equitable and accurate model for predicting graft failure within a multitask learning framework. FERI aims to address the ethical complexities in real-world applications such as transplant allocation and post-transplant success, building on existing research while introducing novel fairness-oriented methodologies.

To summarize, the following are the key highlights of our research:

- **Targeted Fairness**: This study focuses on existing systemic biases affecting liver transplantation, emphasizing gender, age group, and race/ethnicity disparities.

- **Methodological Innovation**: Within a multitask learning framework, we propose a FERI algorithm to balance the subgroup training process through constraints on equitable loss improvements.

- **Clinical Application**: The paper aims to enhance equity in liver transplantation by creating a more accurate and fairer model for predicting graft failure, which is crucial for post-transplant outcomes.

## 2. Data Structure

We utilized a large national dataset of 160,360 liver transplant patients between 1987 and 2018, obtained from the Organ Procurement and Transplantation Network (OPTN) maintained by the United Network for Organ Sharing (UNOS)[37]. This rich dataset includes comprehensive profiles of liver transplant recipients and donors, waiting list registrations, lab results, operative details, and longitudinal outcome data. A key variable we modeled is graft failure, a major determinant of post-transplant prognosis[38].

A common transplantation allocation framework is the Model For End-Stage Liver Disease (MELD) score, which estimates the 90-day mortality rate of patients without transplantation using lab tests for serum creatinine, bilirubin, international normalized ratio blood test, and serum sodium[39]. In contrast to MELD, our study integrates a broader range of 117 features, including 52 recipient variables, such as blood type and BMI, and 65 donor variables. This expanded feature set enabled more detailed predictions of graft failure risk.

The demographic information for liver transplant patients is detailed in **Table 1**, which differentiates between those with and without graft failure. It covers gender, age, race, and ethnicity, highlighting potential disparities. For instance, while gender subgroups were relatively evenly distributed, disparities existed between adults (42.83% of patients) and pediatrics (33.99%) for graft failure. Similarly, differences across racial and ethnic groups exist. This suggests that age, race, and ethnicity contribute to unequal graft failure rates, which we aim to address. Additionally, the table underscores representational imbalances across each subgroup in sensitive attributes.

Table 1. Demographic Information of Graft Failure and Non-Graft Failure in Liver Transplant Patients

|  |  | **Graft Failure** | **Non-Graft Failure** |
|---|---|---|---|
| **Gender** | Male | 41,651 (41.84%) | 57,906 (58.16%) |
|  | Female | 24,761 (42.04%) | 34,131 (57.96%) |
| **Age Group** | Adult | 60,821 (42.83%) | 81,179 (57.17%) |
|  | Pediatric | 5,591 (33.99%) | 10,858 (66.01%) |
| **Race/Ethnicity** | White | 50,064 (43.24%) | 65,719 (56.76%) |
|  | Black | 6,796 (44.77%) | 8,384 (55.23%) |
|  | Hispanic | 7,424 (35.31%) | 13,604 (64.69%) |
|  | Asian | 2,128 (32.95%) | 4,330 (67.05%) |

## 3. Methodology

### 3.1 Tab Transformer Framework

We propose a novel multitask transformer-based model to handle tabular data for predicting graft failure across demographic subgroups in liver transplant patients. The Multitask Tab Transformer model combines the benefits of column embeddings, Transformer layers, and multilayer perceptron (MLPs) to improve performance across multiple tasks[40]. The Multitask Tab Transformer model uses a set $D = \{X, y_1, y_2, ..., y_M\}$ where $X$ is the combined set of categorical and continuous features, $y^m$ is the output label set for task $m$, and $M$ is the total number of tasks.

**Column Embedding Layer:** In the initial stage, categorical features, denoted as $x_{cat} = \{x_1, x_2, ..., x_p\}$, where $p$ represents the number of categorical features is converted into embeddings. Each categorical feature $x_j$, where $j \in \{1, ..., p\}$, is converted to a parametric metric embedding space, accommodating missing values as well. The parametric embedding $E_\phi(x_{cat})$ represents the initial Transformer layer.

**Transformer Layers:** These embeddings serve as inputs to a series of Transformer layers denoted as $f_\theta$, which use a multi-head self-attention mechanism and a position-wise feed-forward layer to capture complex interactions and dependencies between features.

**Multilayer Perceptron:** Following the Transformer layers, both the derived categorical feature embeddings and continuous features, denoted as $x_{cont}$, are layer-normalized and concatenated before being fed into the model. This concatenated vector is passed through a multilayer perceptron, denoted as $g_\psi^m$ for each task $m$ to produce task-specific outputs.

### 3.2 Baseline Model and Limitations

Learning multiple tasks concurrently presents an intricate optimization challenge as balancing competing objectives must be addressed[41]. Typically, the weights of these tasks are either set uniformly across tasks or adjusted through manual tuning[42]. Our baseline model employs an average of the loss from each task for optimization. Given $M$ different tasks, each task has an associated loss function, $L_m(X, y^m)$, which is minimized to optimize all the parameters. This baseline model employs a basic averaged binary cross-entropy loss function defined as $BCE\left(g_\psi^m\left(f_\theta\left(E_\phi(x_{cat})\right), x_{cont}\right), y^m\right)$ across all tasks:

$$Loss_{baseline}(X, Y) = \frac{1}{M} \sum_{m=1}^{M} L_m(X, y^m) \quad (1)$$

here, $M$ is dynamically determined by the sensitive attribute in focus:

- For Gender, $M = 2$ (Male and Female).
- For Age Group, $M = 2$ (Adult and Pediatric).
- For Race/Ethnicity, $M = 4$ (White, Black, Hispanic, and Asian).

The baseline model is prone to issues when tasks have imbalanced data distributions, as is common across subgroups defined by sensitive attributes such as age group, gender, and race/ethnicity. For example, the 'Pediatric' age group may have much less training data than the 'Adult'. In the baseline model, which simply averages the loss, the pediatric task can learn rapidly while adult learning stalls, which results in unequal optimization between tasks. The 'Pediatric' age group loss decreases quickly as it is easier to optimize, while 'Adult' learning progresses slowly.

### 3.3 FERI algorithm: Fairness through Equitable Rate of Improvement in Multitask Learning

We incorporated multitask learning into the Tab Transformer framework to enhance performance and fairness further. Multitask learning is particularly effective for our objectives as it often outperforms single-task models, especially when the tasks are correlated[26]. In our implementation, we utilize a common approach known as hard parameter sharing, which allows the model to share hidden layers across all tasks while maintaining task-specific MLP output layers to produce predictions for each task[43]. Each task in the multitask learning setup is carefully explained based on sensitive demographic attributes, such as gender, age group, and race/ethnicity. This design choice enables us to focus on optimizing the model's performance for each subpopulation as a specific task, aiming to mitigate biases and ensure fairness in prediction across different demographics.

Inspired by foundational principles from existing research, we developed our own FERI algorithm that adjusts task losses to balance loss distribution in multitasking scenarios while extending to mitigate algorithmic biases in model predictions. We adapted the idea of dynamically fine-tuning the subgroup learning rate and efficiently managing training speed. Eventually, FERI adjusts the weights of each subgroup during training to ensure demographic fairness. Rather than solely focusing on balancing the learning rate per task, FERI also aims to robustly mitigate potential biases against sensitive attributes across various subgroups.

At first, we initialize various hyperparameters and attributes such as the number of tasks $M$, the learning rate $\alpha$ and $\beta$ which control the step size in each update during training, a regularization term $\gamma$, and maximum gradient norm. We also consider a stability constant $\epsilon \in \mathbb{R}^+ (= 1 \times 10^{-8})$ to prevent division by zero. Notably, the logits for each task, denoted as $w_{t,m}$, starts at zero. As a result, the *softmax*-transformed weights $z_t$ start by assigning equal importance to every task. The algorithm also starts with the model parameter $\theta_t$ with binary cross-entropy loss function $L_m(\theta_t)$.

Before any training begins, the combined task loss for the entire dataset is given by:

$$Loss_{FERI}(X, Y) = \sum_{m=1}^{M} z_{t,m} \times L_m(\theta_t) \quad (2)$$

here, $z_{t,m} = e^{w_{t,m}} / \sum_{m=1}^{M} e^{w_{t,m}}$ is the weight corresponding to task $m$ at epoch $t$, which is determined by taking the $m^{th}$ entry of the *softmax* transformation applied to the entire vector $w_t = [w_{t,1}, w_{t,2}, \ldots, w_{t,M}]$ for all tasks. For each training epoch $t$, the following steps are undertaken to adjust the loss for the next epoch $t + 1$:

a. **Weight calculation**

For each task $m$, the *softmax* function is applied to the logits of the current epoch to obtain $z_{t,m}$ from the $m^{th}$ entry as $Softmax(w_t)$. This value represents the weight for task $m$ at epoch $t$. Consequently, the vector $z_t = [z_{t,1}, z_{t,2}, \ldots, z_{t,M}]$ aggregates the weights for all $M$ tasks at epoch $t$. To get a better understanding of the model's current performance for each task, an adjusted loss is computed for every task as $\Delta_{t,m} = L_m(\theta_t) + \epsilon$.

b. **Model Parameters Update using Weighted Gradient Descent**

We update the model parameters based on the current epoch weights as:

$$q_{t+1} = q_t - \alpha (\nabla \log \Delta_t) \frac{z_t}{r_t} \quad (3)$$

where $\Delta_t = (\Delta_{t,1}, \Delta_{t,2}, \ldots, \Delta_{t,M})^T$. The $r_t = \sum_{m=1}^{M} z_{t,m}/\Delta_{t,m}$ is used to re-normalize the task weighting back into the probability simplex by dividing a constant $r_t$ per epoch.

c. **Task Weight Logits Update using Weighted Gradient Descent**

Using the updated model parameters $q_{t+1}$, we recompute the task-specific losses using $\Delta_{t+1,m} = L_m(q_{t+1}) + \epsilon$ and collate them in the vector $\Delta_{t+1} = (\Delta_{t+1,1}, \Delta_{t+1,2}, \ldots, \Delta_{t+1,M})^T$. This is crucial as it provides the necessary information to adjust the importance weights or logits of the tasks. Simultaneously, the logits for each task, which guide the weight determination, are also updated. We utilized the following update rule for logits:

$$w_{t+1} = w_t - \beta(\delta_t + \gamma w_t) \quad (4)$$

where $\delta_t = \begin{pmatrix} \nabla^T z_{t,1}(w_t) \\ \vdots \\ \nabla^T z_{t,M}(w_t) \end{pmatrix}^T (log\Delta_t - log\Delta_{t+1})$.

The term $log\Delta_t - log\Delta_{t+1}$ denotes the change in the logarithm of the loss function after the parameter update. This difference indicates how the recent parameter adjustments impacted the overall model performance across tasks. Intuitively, a larger loss change on a task (faster training speed) will induce a smaller weight on this task in the next epoch to promote all tasks being trained at similar speeds. We adaptively adjust task weights by incorporating this into the logits' update.

Initiating with the framework from Equation 2 to 4, the FERI algorithm iteratively adjusts its parameter for each epoch $t$. The weights $z_{t,m}$ and parameters $\theta_t$ are updated to reflect the current epoch. The described algorithm FERI offers a dynamic task weighting mechanism in which each task's importance is continuously updated, and the model parameters are fine-tuned accordingly in each epoch. As training progresses, these logits (and consequently, the weights) will be adjusted to potentially give more importance to tasks that are harder or less importance to tasks that are easier, depending on both individual task losses and the combined task loss from the prior epoch. This adaptive approach ensures the model shifts its emphasis on tasks based on their recent performance, aiming to optimize the combined task loss in subsequent epochs.

**Algorithm 1. Fairness through Equitable Rate of Improvement in Multitask Learning (FERI)**

| | |
|---|---|
| 1 | **Inputs** $D = \{X, y_1, y_2, ..., y_m\}$, number of tasks $M$, Learning rate $\alpha$ and $\beta$, Regularization term $\gamma$, maximum gradient norm is 1. |
| 2 | **Define** sensitive attributes (gender, age group, race/ethnicity) and their subgroups $\{1, ..., M\}$ |
| 3 | **Initialize** $w_{t,m} = 0$ for all $m$ in $M$ |
| 4 | **for** each epoch $t$ in number_epochs do |
| 5 |     **for** each task $m$ in $M$ do |
| 6 |         // Compute Individual Losses and Weights |
| 7 |         calculate $L_m(\theta_t)$ for each task $m$ |
| 8 |         calculate $\Delta_{t,m} = L_m(\theta_t) + \epsilon$ |
| 9 |     **end for** |
| 10 | |
| 11 |     // Model Parameters Update |
| 12 |     **Define** $w_t = [w_{t,1}, w_{t,2}, ..., w_{t,M}]$ |
| 13 |     **Calculate** $z_t = Softmax(w_t) = [z_{t,1}, z_{t,2}, ..., z_{t,M}]$ |
| 14 |     **Calculate** $\Delta_t = (\Delta_{t,1}, \Delta_{t,2}, ..., \Delta_{t,M})^T$ |
| 15 |     **Calculate** $r_t = \sum_{m=1}^{M} \frac{z_{t,m}}{\Delta_{t,m}}$ |
| 16 |     **Updates** $q_{t+1} = q_t - \alpha(\nabla \log \Delta_t) \frac{z_t}{r_t}$ |
| 17 |     // Recalculate Losses with Updated Model |
| 18 |     **for** each task $m$ in $M$ do |
| 19 |         **Recalculate** the loss using $q_{t+1}$ to get $\Delta_{t+1,m} = L_m(q_{t+1}) + \epsilon$ |
| 20 |     **end for** |
| 21 | |
| 22 |     // Update Task Weight Logits |
| 23 |     **Define** $\Delta_{t+1} = (\Delta_{t+1,1}, \Delta_{t+1,2}, ..., \Delta_{t+1,M})^T$ |
| 24 |     **Calculate** $\delta_t = (\nabla z_{t,1}(w_t), \nabla z_{t,2}(w_t), ..., \nabla z_{t,M}(w_t))(log\Delta_t - log\Delta_{t+1})$ |
| 25 |     **Updates** $w_{t+1} = w_t - \beta(\delta_t + \gamma w_t)$ |
| 26 | **end for** |

We summarized **Algorithm 1** which dynamically adjusts the model's weights and parameters, allowing for ongoing optimization and more accurate future forecasts. FERI reduces bias in liver transplantation predictions by dynamically recalibrating weights assigned to different demographic subgroups during training, ensuring no subgroup is favored due to the large sample size. This leads to more equitable outcomes across gender, age group, and race/ethnicity in liver transplant patients.

## 4. Model Evaluation

We assessed the FERI algorithm based on two primary objectives. First, we evaluated its predictive accuracy for liver transplant graft failure across various demographic subgroups, defined by sensitive attributes such as gender, age group, and race/ethnicity. We measured this using the Area Under the Curve of the Receiver Operating Characteristic (AUROC) and the Area Under the Precision-Recall Curve (AUPRC) metrics, comparing the performance of the FERI algorithm against a baseline model for each demographic subgroup. Secondly, we evaluated the model's fairness using demographic parity and equalized odds as fairness metrics. For both training and validation, a five-fold cross-validation approach was employed. Loss plots were analyzed to understand how the convergence rate varies between the FERI and baseline models within each subgroup. Ultimately, the aim is to establish the FERI algorithm's proficiency in balancing both predictive accuracy and fairness when forecasting liver transplant graft failure. For clarity, the fairness metrics used were defined as follows: $\hat{Y}$: the prediction on graft failure made by the model; $Y$: the true label or actual outcome on graft failure; $S$: sensitive attributes.

*Demographic Parity*:

$$\max_{i,j \in A}[P(\hat{Y} = 1|S = i) - P(\hat{Y} = 1|S = j)] \quad (5)$$

*Equalized Odds:*

$$\max_{i,j \in A}[P(\hat{Y} = 1|S = i, Y = 1) - P(\hat{Y} = 1|S = j, Y = 1)]$$
$$+ \max_{i,j \in A}[P(\hat{Y} = 1|S = i, Y = 0) - P(\hat{Y} = 1|S = j, Y = 0)] \quad (6)$$

## 5. Experiments

In the following experiments, we monitor training loss rates across demographic subgroups for both the FERI and baseline models to minimize disparities during training. FERI's efficacy in addressing fairness for liver transplantation was evaluated by adjusting subgroup losses and comparing performance to baseline models on demographic parity and equalized odds. FERI's accuracy against the baseline was assessed using AUROC and AUPRC for subgroups within sensitive attributes to determine if FERI can improve fairness without sacrificing evaluation accuracy for predicting liver transplant graft failures.

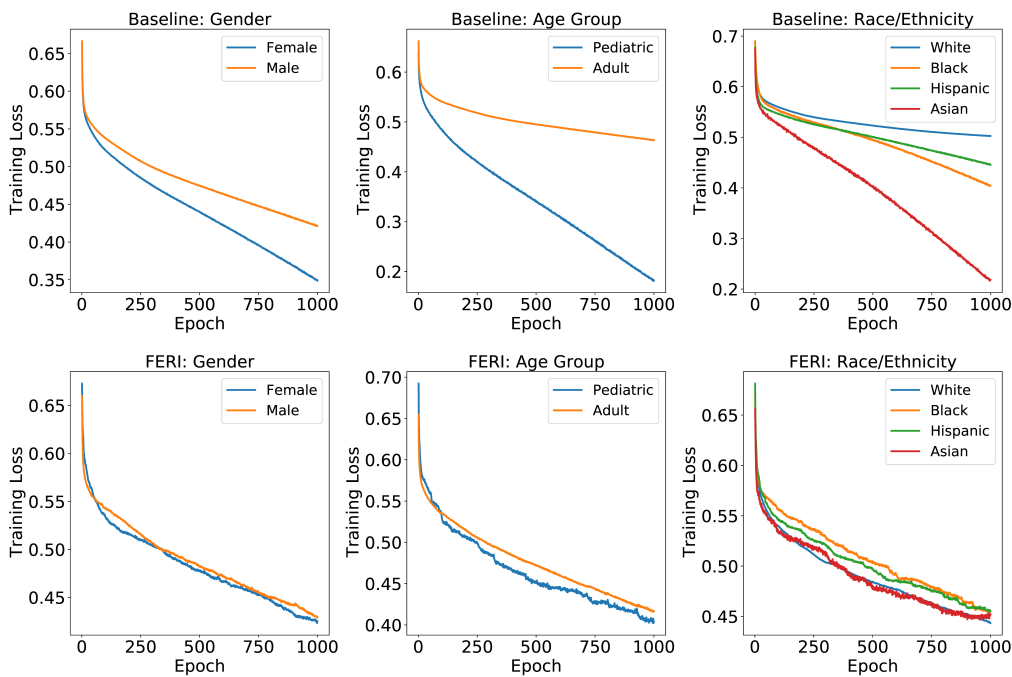

**Figure 1**. Training Losses Over Epochs for Different Sensitive Attributes. The figure represents a comparative analysis of the baseline and FERI's loss progression across epochs, categorized by settings: Gender, Age Group, and Race/Ethnicity. X-axis: Epochs, Y-axis: Loss Value.

### 5.1 Loss Progression Analysis: FERI vs. Baseline Across Subgroups

We comprehensively assessed the FERI algorithm's behavior during training, evaluating its ability to adapt and be unbiased across diverse demographic groups. We compared the FERI algorithm to baseline models to evaluate its improvements in adaptability and fairness.

To monitor the trajectory of training loss across epochs, we observed changes in losses from one epoch to the next, specifically focusing on the demographic subgroups of gender (Male, Female), age group (Adult, Pediatric), and race/ethnicity (White, Black, Hispanic, Asian). Furthermore, hyperparameter tuning played a crucial role in this experiment. We varied the regularization coefficient ($\gamma$) from 1e-9 to 1e-2 and adjusted the learning rate for task logits $\alpha$ between 0.05 and 0.15. Using the optimized hyperparameters ($\gamma$ is 1e-6 for both Gender and Age, and 1e-7 for Race/Ethnicity, along with $\alpha$ is 0.11 for Gender, 0.10 for Age, and 0.08 for Race/Ethnicity), **Figure 1** provides a clear illustration of the effectiveness of the FERI algorithm. The algorithm achieved a uniform loss-reduction trajectory across various demographic subgroups. For example, both the male and female subgroups reached a loss value of 0.5 by approximately the 300-epoch benchmark. This consistency emphasizes the model's capability to equitably enhance performance across subgroups. The FERI algorithm's loss curves were curlier than the baseline because FERI updates each epoch's losses based on the previous epoch's outcomes. This step-by-step adjustment process allows FERI to incrementally refine the losses through continuous monitoring.

### 5.2 Comparative Fairness Analysis: FERI vs. Baseline

We evaluated the effectiveness of our proposed FERI algorithm in addressing fairness issues against baseline models through an ablation study. We performed a five-fold cross-validation method to split the data into 60% for training, 20% for validation, and 20% for testing.

**Table 2**. Disparity Analysis: Demographic Parity and Equalized Odds Comparisons between Baseline and FERI. Each entry provides fairness metrics value from a five-fold cross-validation. FERI algorithm performance on demographic parity and equalized odds are boldfaced and highlighted in green for each fold. A smaller value in our fairness metrics indicates reduced unfairness.

| Sensitive attribute | Fold Number | Baseline | FERI | Baseline | FERI |
|---|---|---|---|---|---|
| | | **Demographic Parity** | | **Equalized Odds** | |
| **Gender** | 1 | 0.0062 | 0.0013 | 0.0312 | 0.0246 |
| | 2 | 0.0084 | 0.0002 | 0.0148 | 0.0055 |
| | 3 | 0.0070 | 0.0042 | 0.0287 | 0.0225 |
| | 4 | 0.0012 | 0.0005 | 0.0198 | 0.0118 |
| | 5 | 0.0001 | 0.0005 | 0.0278 | 0.0183 |
| | **Mean ± SD** | 0.0046 (±0.0037) | **0.0013 (±0.0017)** | 0.0245 (±0.0069) | **0.0165 (±0.0079)** |
| **Age Group** | 1 | 0.0681 | 0.0501 | 0.0699 | 0.0633 |
| | 2 | 0.0774 | 0.0369 | 0.0830 | 0.0158 |
| | 3 | 0.0885 | 0.0711 | 0.0972 | 0.0704 |
| | 4 | 0.0833 | 0.0473 | 0.0951 | 0.0460 |
| | 5 | 0.0710 | 0.0576 | 0.0714 | 0.0524 |
| | **Mean ± SD** | 0.0777 (±0.0084) | **0.0526 (±0.0127)** | 0.0833 (±0.0128) | **0.0496 (±0.0211)** |
| **Race/Ethnicity** | 1 | 0.1190 | 0.0887 | 0.2035 | 0.1405 |
| | 2 | 0.1253 | 0.1170 | 0.2039 | 0.2014 |
| | 3 | 0.1175 | 0.1087 | 0.1809 | 0.1777 |
| | 4 | 0.1130 | 0.0984 | 0.1984 | 0.1710 |
| | 5 | 0.1033 | 0.1010 | 0.1685 | 0.1665 |
| | **Mean ± SD** | 0.1156 (±0.0081) | **0.1028 (±0.0107)** | 0.1910 (±0.0157) | **0.1714 (±0.0219)** |

In **Table 2**, it shows the comparison across sensitive attributes of gender, age group, and race/ethnicity. This emphasizes the FERI algorithm's superior performance in reducing disparities across sensitive attributes versus baselines. For example, when considering demographic parity with respect to gender, the FERI algorithm showed an obvious improvement. The baseline model had a demographic parity score of 0.0046 with a standard deviation of ±0.0037. In contrast, the FERI algorithm scored 0.0013 with a standard deviation of ±0.0017. The reduction was derived by taking the difference between the baseline model value and the FERI algorithm value, and then dividing the result by the baseline model value. This represents a reduction in disparity of approximately 71.74%. Similarly, in the age group category, the equalized odds metric saw a decline from the baseline model's score of 0.0833 (with a standard deviation of ±0.0128) to the FERI algorithm's score of 0.0496 (with a standard deviation of ±0.0211). This shows an approximate reduction of 40.46% in the disparity. The FERI algorithm also displayed improvements for race and ethnicity, underscoring its efficacy in promoting fairness across all sensitive attributes. In conclusion, the FERI algorithm consistently outperformed the baseline, indicating reduced unfairness across all examined attributes.

### 5.3 Accuracy Sacrifice Evaluation: FERI vs. Baseline

For a robust evaluation, we adopted a five-fold cross-validation approach for evaluating the accuracy using both AUROC and AUPRC metrics. The values presented in **Figure 2** and **Figure 3** were averages from these five folds for each subgroup within the specified sensitive attributes. We presented a detailed analysis that compares the predictive performance of our FERI algorithm against the baseline model. Remarkably, the FERI algorithm exhibited AUROC and AUPRC scores that closely align with those of the baseline model across all analyzed subgroups. The notation |Δ| in the figures represents the absolute difference between the FERI and baseline values, highlighting a minimal trade-off. Specifically, the AUROC values for both models consistently exceeded the 0.7 threshold and AUPRC values are above 0.6. They provided robust evidence that our FERI algorithm succeeded in improving fairness metrics without sacrificing predictive accuracy. Meanwhile, it also underlined the model's capabilities in ensuring an equitable prediction of graft failure outcomes in liver transplantation across various demographic subgroups.

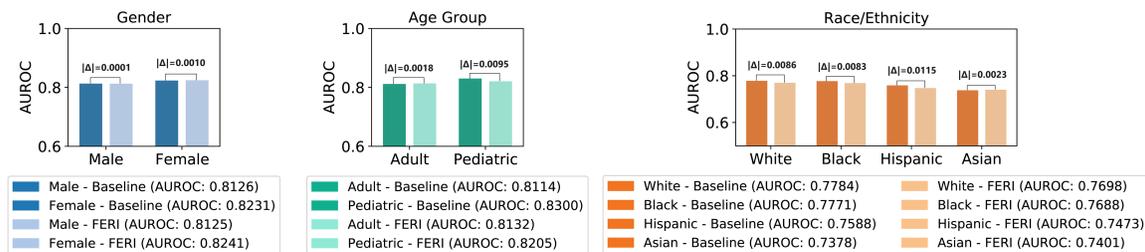

**Figure 2.** Evaluation of Accuracy Trade-offs in Baseline and FERI Using AUROC Metrics for Subgroup

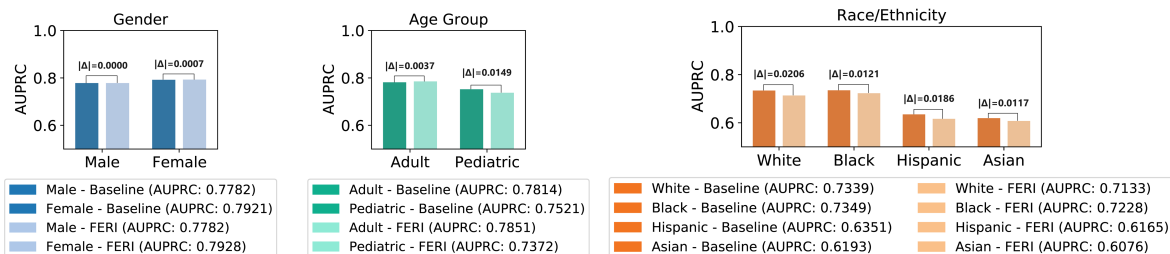

**Figure 3.** Evaluation of Accuracy Trade-offs in Baseline and FERI Using AUPRC Metrics for Subgroups

Through all these three experiments, FERI consistently outperformed the baseline models, demonstrating its effectiveness in balancing both predictive accuracy and fairness in predicting liver transplant graft failure across all sensitive attributes.

### 6. Discussion

Our multitask-based techniques equalize and optimize the training rates across all subgroups in different demographic populations, thereby improving fairness by mitigating disparate training speeds. However, it's crucial to note that FERI is sensitive to hyperparameters such as learning rate and weight decay. Fine-tuning is often required for FERI to achieve optimal performance on specific datasets. Future work will extend beyond algorithmic validation, emphasizing real-world settings and focusing on diverse patient populations. We aim to develop an adaptive mechanism to counteract biases arising from imbalanced subgroup representation within the dataset. The goal is to optimize hyperparameters for generalizability and develop techniques to improve robustness to skewed subgroup distributions. Moving forward, our focus will include non-technical factors such as the varying prevalence of diseases among demographic groups to enhance our algorithm's applicability across diverse clinical settings. We will refine our approach to better account for the above factors in future algorithm improvement. Overall, developing accurate and unbiased predictive tools is essential to reduce healthcare disparities and ensure clinically meaningful predictions for all patients. This work provides a valuable framework for future efforts to integrate both accuracy and fairness objectives.

## 7. Conclusion

In conclusion, the FERI algorithm provides an innovative approach to fair predictive modeling in healthcare. By utilizing multitask learning and dynamically adjusting loss function weights, FERI achieved comparable overall accuracy to baseline models based on AUROC and AUPRC metrics. Critically, it also demonstrated substantially improved fairness by equalizing improvement rates across demographic subgroups.

## 8. Acknowledgment

XJ is a CPRIT Scholar in Cancer Research (RR180012), and he was supported in part by the Christopher Sarofim Family Professorship, the UT Stars award, UTHealth startup, the National Institute of Health (NIH) under award number R01AG066749, R01AG066749-03S1, R01LM013712, R01LM014520, R01AG082721, R01AG066749, U01AG079847, U24LM013755, U01TR002062, U01CA274576, U54HG012510 and the National Science Foundation (NSF) #2124789